\documentclass[10pt,twocolumn,letterpaper]{article}

\usepackage{cvpr}              

\usepackage{graphicx}
\usepackage{amsmath}
\usepackage{amssymb}
\usepackage{booktabs}

\usepackage{wrapfig}

\usepackage{diagbox}
\usepackage{xcolor}
\usepackage{colortbl}
\usepackage{pifont}
\newcommand{\cmark}{\ding{51}}%
\newcommand{\xmark}{\ding{55}}%

\usepackage{marvosym}
\usepackage{algorithm}
\usepackage{algpseudocode}

\usepackage{diagbox}
\usepackage{multirow}

\usepackage[accsupp]{axessibility}

\definecolor{cvprblue}{rgb}{0.21,0.49,0.74}
\usepackage[pagebackref,breaklinks,colorlinks,allcolors=cvprblue]{hyperref}

\def\paperID{2} 
\def\confName{CVPR}
\def\confYear{2025}

\title{CityGen: Infinite and Controllable City Layout Generation}

\author{
Jie Deng\textsuperscript{1,$*$} \quad
Wenhao Chai\textsuperscript{2,$*$, $\dagger$} \quad
Jianshu Guo\textsuperscript{1,$*$} \quad
Qixuan Huang\textsuperscript{1} \quad
Junsheng Huang\textsuperscript{1} \\
Wenhao Hu\textsuperscript{1} \quad
Shengyu Hao\textsuperscript{1} \quad
Jenq-Neng Hwang\textsuperscript{2} \quad
Gaoang Wang\textsuperscript{1,\Letter} \\
[2mm]
\textsuperscript{1}~Zhejiang University \quad
\textsuperscript{2}~University of Washington \\
[2mm]
\normalsize{
\textsuperscript{$*$}~Equal contribution \quad \textsuperscript{$\dagger$}~Project lead \quad \textsuperscript{\Letter}~Corresponding author} \\
[2mm]
}

\begin{document}
\maketitle
\begin{abstract}
The recent surge in interest in city layout generation underscores its significance in urban planning and smart city development.  The task involves procedurally or automatically generating spatial arrangements for urban elements such as roads, buildings, water, and vegetation. Previous methods, whether procedural modeling or deep learning-based approaches like VAEs and GANs, rely on complex priors, expert guidance, or initial layouts, and often lack diversity and interactivity. In this paper, we present CityGen, an end-to-end framework for infinite, diverse, and controllable city layout generation. Our framework introduces an infinite expansion module to extend local layouts to city-scale layouts and a multi-scale refinement module to upsample and refine them. We also designed a user-friendly control scheme, allowing users to guide generation through simple sketching. Additionally, we convert the 2D layout to 3D by synthesizing a height field, facilitating downstream applications. Extensive experiments demonstrate CityGen's state-of-the-art performance across various metrics, making it suitable for a wide range of downstream applications.
\end{abstract}
    
\section{Introduction}

Virtual city generation~\cite{shang2024urbanworld,lin2023infinicity, xie2023citydreamer, cityengine, gao2024embodiedcity, wu2024metaurban, deng2024citycraft} is a rising task in computer vision. Creating realistic and diverse digital copies of urban environments enables the exploration of various architectural designs and infrastructure scenarios. City generation involves a complex blend of sub-tasks critical for crafting comprehensive and realistic urban environments. For example: city layout generation~\cite{groenewegen2009procedural,he2023globalmapper,he2024coho}, urban planning~\cite{parish2001procedural,chu2019neural,xu2021blockplanner}, traffic simulation~\cite{hu2018urban, lu2020impact,li2024chatsumo}, and \etc. Among these components, the creation of city layouts is particularly notable, representing both the foundational and one of the most critical steps. 

Generating city layouts faces significant challenges, particularly in scalability, efficiency, reliance on prior knowledge, and accurately capturing urban complexities. During the past years, various attempts have been to generate realistic city layouts, including procedural modeling~\cite{groenewegen2009procedural, lipp2011interactive}, which progressively creates city layouts based on predefined rules and algorithms, and image-based modeling~\cite{aliaga2008interactive,fan2014structure,henricsson1996automated,vezhnevets2007interactive}, which generate layouts using images, such as aerial, satellite, and street images. While these techniques successfully generate city layouts, their reliance on complex prior knowledge and limitations in scalability and efficiency leads to significant drawbacks. Researchers have employed technologies such as Variational Auto Encoders (VAEs)~\cite{jyothi2019layoutvae}, Generative Adversarial Networks (GANs)~\cite{li2019generating}, transformers~\cite{kong2022blt, wen2022layoutransformer}, and diffusion models~\cite{inoue2023layoutdm} for layout generation in various domains. However, these methods predominantly rely on axis-aligned bounding boxes to represent object positions, which fall short in capturing the complexities of urban environments. While alternative strategies~\cite{xu2021blockplanner, he2023globalmapper} that utilize rotated boxes for buildings allow for varied orientations, they impose shape constraints that curtail their utility. Recent advancements~\cite{lin2023infinicity, xie2023citydreamer} have shifted towards representing city layouts as semantic masks to improve urban modeling details. Despite these efforts, achieving diversity and controllability in generated layouts remains a significant hurdle, underscoring the need for continued innovation in this field.  

To address these issues, we propose CityGen, an end-to-end framework to generate \textbf{infinite}, \textbf{diverse}, and \textbf{controllable} city layouts. Besides generating 2D layouts, we utilize an efficient representation for 3D city layout which only consists of a semantic field and a height field~\cite{chen2023scenedreamer}.
To generate an infinite-scale semantic field, we first employ a diffusion model~\cite{ho2020denoising} to create a local semantic block and then construct an infinite expansion model to extend it into a global semantic field. It is noteworthy to highlight that we train the diffusion model in the one-hot space, as we represent the semantic map as a one-hot vector. This training approach exhibits robust performance when compared with the conventional RGB image space. After that, we develop a multi-scale refinement framework that progressively upsamples low-resolution local semantic blocks and combines them to form a coherent high-resolution global semantic field. The multi-scale refinement framework seeks to integrate broad global semantics with fine-grained details at lower scales, effectively minimizing the computational costs of training and sampling high-resolution models. 

Additionally, we design a user-friendly scheme to facilitate user control over the generation of layouts. Given any user sketches of a city layout, our model can produce high-quality, diverse samples devoted to user customization. 
To synthesize a height field for a given semantic field, we separate classes and model heights for each class according to elevation patterns we observe in the real world. We further segment buildings into instances to enable precise and diverse height assignments for each building instance. We combine these submodules to provide realistic, infinite, and diverse city layouts. Our main contributions can be summarized as follows:

\begin{itemize}
\item We propose CityGen, an end-to-end framework to generate infinite, controllable and diverse city layouts. 
\item We propose a multi-scale refinement framework that progressively upscales and refines the generated layouts, allowing us to train models at the smallest scale, reducing computational and data requirements while ensuring diversity and quality in the results.
\item We design a user-friendly scheme that allows users to guide the generation of customized layouts tailored to their designs through simple sketching interactions.
\end{itemize}
\section{Related Work}
\subsection{3D Scene Generation}
3D Scene Generation focuses on the creation of efficient 3D representations of scenes, with a primary categorization into indoor~\cite{devries2021unconstrained,gao2023scenehgn,ritchie2019fast,shi20223d} and outdoor scenes~\cite{chen2023sd,chai2023persistent}. Outdoor scenes further encompass natural~\cite{chan2022efficient,devries2021unconstrained,hao2021gancraft, chai2023persistent} and urban~\cite{shen2022sgam, lin2023infinicity, xie2023citydreamer} environments. Scenes can be represented explicitly using voxels~\cite{kim20133d,li2023voxformer}, point clouds~\cite{chen2022stpls3d,hu2021towards}, occupancy~\cite{tong2023scene,wang2023panoocc}, or implicitly using Neural Radiance Fields (NeRF)~\cite{tancik2022block,turki2022mega,xiangli2022bungeenerf}. Recently, notable advancements have been made in the generation of infinitely extending natural~\cite{chen2023scenedreamer,liu2021infinite,li2022infinitenature} and urban scenes~\cite{lin2023infinicity,xie2023citydreamer}. More recently, InfiniCity~\cite{lin2023infinicity} generates BEV map and uses 3D octree-based voxel completion for scene construction, CityDreamer~\cite{xie2023citydreamer} focuses on rendering after city layout creation, and BlockFusion~\cite{wu2024blockfusion} applies diffusion-based model to generate 3D scenes
as unit blocks. 
However, those works have limited considerations for diversity and user-control in 3D city layout generation, which are further explored in this paper.
    
\subsection{Layout Generation}
The development of layout generation techniques has evolved significantly, transitioning from embedding design principles in manually-defined energy functions to adopting sophisticated machine learning models. Initially, traditional methods such as~\cite{o2015designscape, o2014learning} depended on manually crafted energy functions. The field then saw substantial advancements with the introduction of machine learning-based approaches: LayoutVAE~\cite{jyothi2019layoutvae} utilized the VAE framework for generating stochastic scene layouts; LayoutGAN~\cite{li2019layoutgan} applied a GAN to create layouts by capturing the geometric relationships among various 2D elements; LayoutTransformer~\cite{wen2022layoutransformer} leveraged transformer networks to transform input scene graphs into layout representations; and LayoutDM~\cite{inoue2023layoutdm} employed a discrete state-space diffusion model for progressively synthesizing noiseless layouts from structured layout data in a discrete format. While these methods have proven effective for generating layouts across a range of objects and scenes, they predominantly utilize primitive shapes such as rectangles and triangles to denote object locations. This reliance limits their applicability in urban scene generation, where objects often exhibit a wide array of shapes, orientations, and intricate interrelations.

\subsection{Layout Generation for City}
The generation of city layouts involves multifaceted considerations within urban planning, encompassing aspects such as road networks, land use zoning, and the precise placement of buildings. These layouts can be created through a spectrum of methods, ranging from rule-based manual design~\cite{bacon1976design,calthorpe2001regional}, semi-automated processes via procedural generation~\cite{ghorbanian2019procedural,groenewegen2009procedural} using tools like CityEngine~\cite{cityengine} and Unreal Engine~\cite{unrealengine}, to fully automated techniques utilizing deep learning methodologies~\cite{xu2021blockplanner, he2023globalmapper}. However, their representation of layouts as box-shaped geometry limits their freedom to capture the complex shapes of objects in city scenes. More recently, ~\cite{lin2023infinicity,xie2023citydreamer} represent object layouts as semantic masks, but these methods have limited considerations for controllability, therefore lacking coherence for user-controlled long-term generation. In comparison to established techniques, our approach addresses these issues, offering a more diversified and user-controlled process for high-quality city layout generation.

\begin{figure*}[t]
    \centering
    \includegraphics[width=\textwidth]{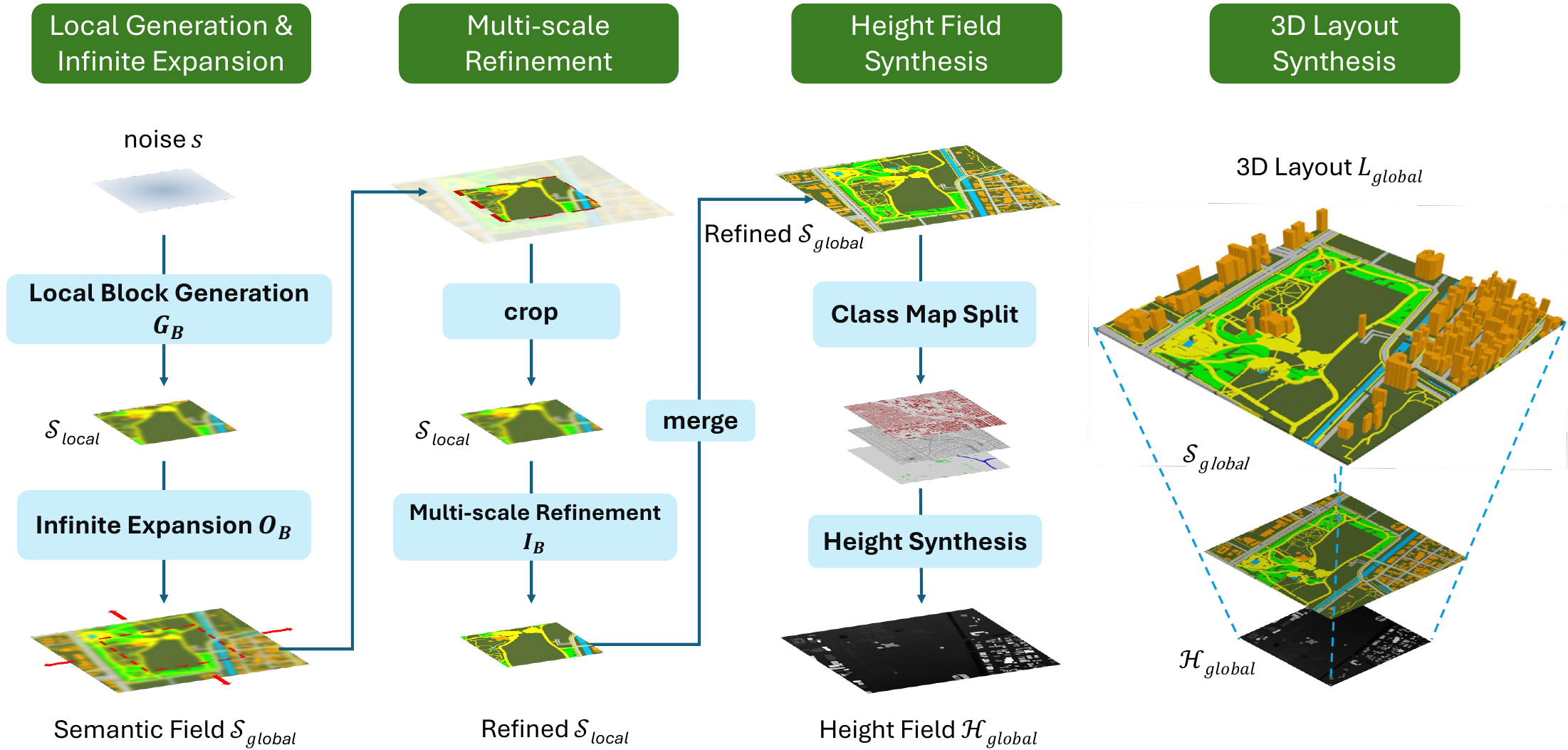}
    \caption{\textbf{CityGen Framework.} In the initial step, we sample a local block from Gaussian noise and extend it infinitely through the infinite expansion module. Subsequently, we iteratively refine and upsample each local region using the multi-scale refinement module to get a refined global semantic field. After that, we split the semantic field and synthesis height field for each class. Finally, we integrate the semantic field with the height field to synthesize the 3D semantic layout. }
    \label{fig:framework}
\end{figure*}
\section{Methodology}
In this section, we outline our approach to generating infinite and controllable 3D city layouts. We begin by generating a local semantic field, and progressively expand it using the infinite expansion module. To refine details and upsample the results, we apply the multi-scale refinement module. We then synthesize the height field, $\mathcal{H}$. Finally, we combine the semantic and height fields to construct 3D layouts. We show the overall framework in~\cref{fig:framework}.

\subsection{Semantic Field Generation}
The semantic field $\mathcal{S}$ represents the semantic labels (terrain, vegetation, building, traffic roads, rail, and water) of each position, which is a one-hot coding map of resolution $H \times W \times C_s$, where $H, W$ are height and width, and $C_s$ is the number of classes.  Generating the full semantic field of a city at once leads to intractable complexity and insufficient detail. Thus, we initiate the semantic field by generating a local block, then use an infinite expansion model to extends it to user-defined scale and use a multi-scale refinement module to upsample and refine each local block to the highest resolution.

\paragraph{Local Block Generation.}
One of the main challenges in generating city layouts is scalability. Most generative models produce fixed-size images, which limits the area they can represent. For instance, with a pixel-to-meter ratio of 0.5, a 512$^2$ image covers only 256$^2$ m². To generate detailed images over large areas, the model must be capable of continuously expanding the layout. Given the strong inpainting and outpainting capabilities of diffusion models, we train one-hot diffusion models~\cite{ho2020denoising}, $\mathrm{G}_B(\cdot)$, as foundational block generation models. Specifically, we generate a semantic field $\mathcal{S}$ with $C_s$ channels. During training, $\mathrm{G}_B(\cdot)$ serves as the denoising network, optimized by minimizing the objective in~\cref{block generation obj}.
\begin{equation}
    \min_{\theta_{\mathrm{G}_B}} 
    \mathbb{E}_{\epsilon \sim \mathcal{N}(0, I), t \sim \mathcal{U}(1, T)}
    \| \epsilon -  \mathrm{G}_B (s_t, t)\|_2^2,
    \label{block generation obj}
\end{equation}
where $\epsilon \sim \mathcal{N}(0, I)$ is random noise sampled from a Gaussian distribution, $s_t \in \mathbb{R}^{H \times W \times C_s}$ is the noisy semantic field at timestep $t$, and $\theta_{\mathrm{G}_B}$ are the parameters of the denoising U-Net~\cite{ronneberger2015u} $\mathrm{G}_B$. Afterward, the denoised $s$ is randomly sampled during inference, and we convert it back to a one-hot semantic field $\mathcal{S}$ using the $argmax(\cdot)$ function. We train three block generation models at scales (128$^2$, 256$^2$, 512$^2$), with all images down-sampled to 128$^2$ during training.

\paragraph{Infinite Expansion.}
The Infinite Expansion module is designed to outpaint a local block to any user-defined scale. It is built on the block generation model trained with images at the largest scale (512$^2$) to prevent a common issue in auto-regressive models—repeated content in long-term generation. Since samples at the largest scale contain the richest semantic information, this helps produce diverse results. The outpainting model $\mathrm{O}_{B}(\cdot)$, a diffusion model with a similar architecture and training objective to $\mathrm{G}_{B}(\cdot)$. It takes an additional outpainting target mask as input. The binary mask $m_{out}$, with $m_{out} = 1$ for known regions and $m_{out} = 0$ for target positions, guides the model to predict unknown regions during training. The training objective for the outpainting model $\mathrm{O}_B(\cdot)$ is defined as:
\begin{equation}
    \min_{\theta_{\mathrm{O}_B}} 
    \mathbb{E}_{\epsilon \sim \mathcal{N}(0, I), t \sim \mathcal{U}(1, T)}
    (\| \epsilon -  \mathrm{O}_B (s_t, t, m_{out})\|_2^2) \odot  (1 - m_{out}),
    \label{outpainting obj}
\end{equation}
where $\theta_{\mathrm{O}_B}$ are the parameters of $\mathrm{O}_B$, and $\odot$ denotes element-wise multiplication, we multiply the loss by $(1 - m_{out})$ so that only the unknown regions contribute to the loss during training. During inference, by sliding the outpainting mask $m_{out}$ in different directions with varying overlap ratios, we progressively expand the local semantic field into the desired global size. To model overlapping regions in the sliding window, we design the outpainting mask with rectangular unknown areas along the boundaries. To ensure smooth transitions between known and outpainted regions, we merge these areas at each denoising step using~\cref{denoising_merge} \cite{lugmayr2022repaint,avrahami2022blended}.
\begin{equation}
    s_{t} = s_{t, ori} \odot m_{out} + s_{t, pred} \odot (1 - m_{out}).
    \label{denoising_merge}
\end{equation}
where $s_{t, ori}$ is the noisy version of the original image and $s_{t, pred}$ is the predicted clean image at timestep $t$.
\paragraph{Multi-scale Refinement Module.}
Since all block generation models are trained on images downsampled to $128^2$, the generated images must be upsampled to their original resolution. This downsampling during training leads to a loss of details, and direct upsampling introduces artifacts. To resolve this, we designed a multi-scale refinement module with two inpainting models $\mathrm{I}_{B}(\cdot)$ based on block generation models at smaller scales ($128^2, 256^2$). These inpainting models are trained similarly to the outpainting model, but with a randomly applied mask, where any pixel may be masked, with a ratio $\sigma \in (0.1, 0.9)$, as artifacts can occur anywhere.

The core idea is that the inpainting model, trained at smaller scales, captures finer details and corrects artifacts from direct upsampling. For a sampled image, we first upsample it by a factor of 2, then crop regions at the previous scale and use the inpainting model to fix artifacts within these patches. Artifacts generally fall into two categories: (1) unclear object boundaries and (2) isolated irregular pixels, like a single water/vegetation pixel in a traffic road. We mask object edges and irregular pixels as unknown regions before applying the inpainting model. Since the layout is a discrete semantic mask with class labels at each pixel, detecting edges and small irregular instances is straightforward after isolating all instances.

\paragraph{User-customization via Inpainting.} 
The inpainting models enable natural, intuitive user interactions. Given a user sketch, we crop it into patches at the inpainting model scales, convert each patch into an incomplete semantic field $\mathcal{S}_{local}^{'}$, and feed it into the corresponding inpainting model to generate a complete semantic layout $\mathcal{S}_{local}$.

\begin{figure*}[t]
    \centering
    \includegraphics[width=\linewidth]{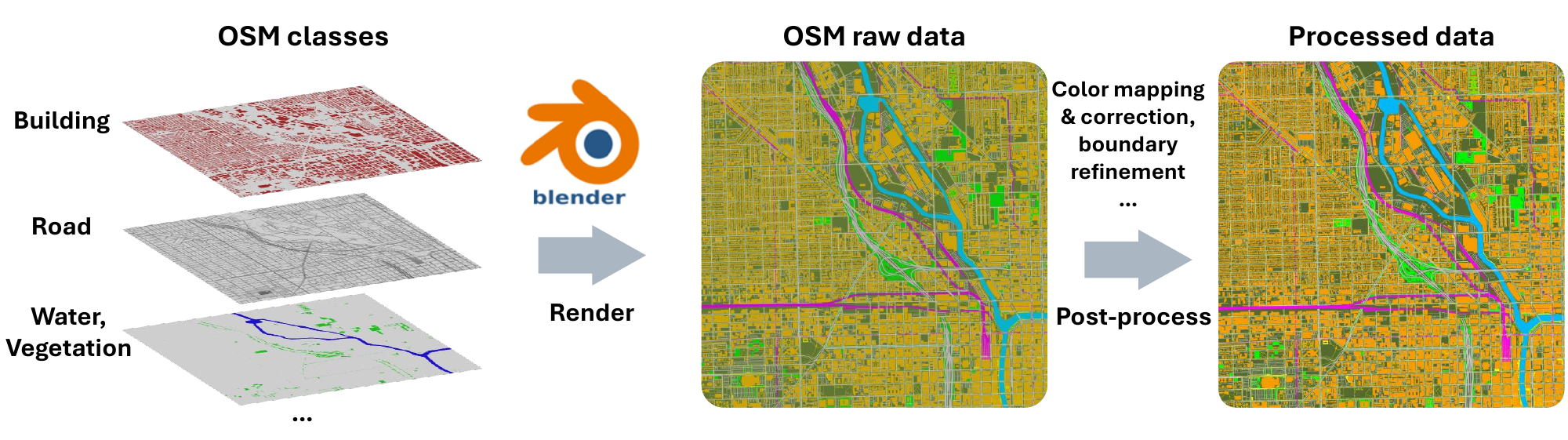}
    \caption{\textbf{OSM data pre-processing pipeline}. We first populate 3D city models for real-world cities with various geospatial elements like buildings, roads, bodies of water, and vegetation. Subsequently, we render the 3D city models with Blender~\cite{Blender}. Then, we do boundary refinement and color correction.}
    \label{fig:osm processing}
\end{figure*}

\subsection{Height Field Synthesis}
\paragraph{BEV Layout Representation.} 
We aim to efficiently and expressively represent 3D city layouts using a bird's-eye-view (BEV) scene representation consisting of a semantic field $\mathcal{S}$ and a height field $\mathcal{H}$, instead of voxel or point cloud-based methods. Given a semantic map $S$, we generate a height field for each class. While it is theoretically possible to generate both fields simultaneously using a multi-task network as in~\cite{xie2023citydreamer, lin2023infinicity}, we found this approach only ensures local consistency between the fields. As the autoregressive process continues, inconsistencies arise. Additionally, due to noisy OpenStreetMap (OSM)~\cite{OpenStreetMap} data, generated height fields often contain irregularities, especially for building rooftops that should be flat. This becomes more evident in 3D visualization. Moreover, since building heights vary significantly while other types (e.g., terrain, vegetation, roads) have minimal elevation changes, modeling all heights with the same distribution is unsuitable. Therefore, we adopt a rule-based approach to synthesize height fields for each class separately. We create a binary mask $M_i$ for each class $C_i$, where $M_i = 1$ for indices where $S = i$ and $M_i = 0$ elsewhere. Terrain, vegetation, and water surfaces exhibit uniform height variations, so we use different simplex noise~\cite{olano2004real} to synthesize their height fields. For roads, footpaths, and rails, which are flat due to human design, we assign fixed heights and overlay them above terrain or vegetation. Buildings, with greater height variations, are processed by detecting connected components and assigning a unique height to each instance. Building heights are determined by randomly sampling from Gaussian distributions specific to building types (e.g., townhouse, apartment, office building), each with distinct mean and variance parameters.

\subsection{3D Layout Construction}
Given a semantic field $\mathcal{S}$ representing the semantic label of the highest surface point in the vertical direction, and the corresponding height field $\mathcal{H}$ representing the highest elevations for all points, we can construct a 3D city layout by lifting each point to its corresponding height as shown in~\cref{fig:framework}. After that, the 2D layout can be converted a to 3D semantic layout by sampling along the vertical axis with a user-specified resolution.
\section{Experiments}
\begin{table*}[t]
\centering

\caption{Functionality comparisons with related layout generation methods.}
\resizebox{0.96\linewidth}{!}{
    \begin{tabular}{l|c|c|c|c|c}
    \toprule
     Method & Prior-Free (If not, prior type) & Object Types & Building Shapes & User Control & Scalability\\
    \midrule
    LayoutVAE~\cite{jyothi2019layoutvae} & label set & general & axis-aligned box only & \xmark & \xmark \\ 	  
    LayoutTransformer~\cite{wen2022layoutransformer} & scene graph & general & axis-aligned box only & \xmark & \xmark \\
    GlobalMapper~\cite{he2023globalmapper} & road networks & building + road & arbitrary but constrained & \cmark & \cmark \\
    BlockPlanner~\cite{xu2021blockplanner} & \cmark & building + road & box only &  \xmark  & \cmark \\
    Infinicity~\cite{lin2023infinicity} & \cmark & 4 types & arbitrary & \xmark  & \cmark \\
    CityDreamer~\cite{xie2023citydreamer} & \cmark & 6 types & arbitrary &  \xmark  & \cmark \\
    \midrule
    \textbf{CityGen~(ours)} & \cmark & \textbf{7 types} & \textbf{arbitrary} &  \cmark & \cmark\\
    \bottomrule
    \end{tabular}
}
\label{table:comparison_functionality}
\end{table*}

\subsection{Dataset} 
Our training data for the block generation comes from the OpenStreetMap~(OSM)~\cite{OpenStreetMap}, a free open geographic database updated and maintained by a community of volunteers via open collaboration.
We first populate 3D city models for real-world cities with various geospatial elements like buildings, roads, bodies of water, and vegetation. Subsequently, we render the 3D city models with Blender~\cite{Blender}.
After that, we set the camera configurations to get orthogonal aerial projections of the terrain models of size $8192^2$ with a pixel-to-meter ratio of 2:1 to represent an area of $4096^2$ m$^2$ in the real world.
During the rasterization process, the RGB values of 2D views undergo subtle alterations, particularly at the boundaries between distinct object types, attributable to rendering artifacts. To resolve these artifacts, we develop a post-processing pipeline that consists of boundary refinement and color correction. This procedural requirement is essential due to the pivotal role of colors in representing class labels within our framework. Any misalignment in colors can potentially lead to inaccurate representations of class information, introducing a risk of false indications. Subsequently, we crop these high-resolution projections to generate patches of optimal sizes including $128^2$, $256^2$, and $512^2$ to facilitate the training of models at all scales. We demonstrate the data processing pipeline in~\cref{fig:osm processing}. After cropping, we have in total 268556 $512^2$ images, 1074244 $256^2$ images, and 4293888 $128^2$ images.

\subsection{Implementation Details}
We train three block generation models at scales of $128^2$, $256^2$, and $512^2$, with all training images down-sampled to $128^2$ using nearest neighbor interpolation. The outpainting model in the infinite expansion module and the inpainting models in the multi-scale refinement module are based on block generation models at $512^2$, $256^2$, and $128^2$ resolutions, respectively, and further refined with additional data corresponding to their scales. The outpainting and inpainting models are initialized with the weights of the block generation models trained at corresponding image scales and we add $C_{s}+1$ extra input channels initialized with zero weights to the first $conv\_in$ layer of the UNet in the channel dimension, such that it takes $Concat(m, m \odot S_{t}, S_{t} )$ as input, where $Concat(\cdot)$ stands for concatenation in the channel dimension and $m$ is an inpainting or outpainting mask. We use Adam~\cite{kingma2014adam} as the optimizer for training all models. We use a learning rate of $1 \times 10^{-4}$ for all block generation models and $1 \times 10^{-5}$ for all inpainting and outpainting models.

\subsection{Functionality Comparisons}
Significant exploration has been conducted in traditional urban layout generation, however, most works have focused on relatively coarse tasks, such as generating road networks~\cite{karagiorgou2012vehicle, song2019townsim, chen2008interactive, chu2019neural} or generating building distributions only~\cite{showkatbakhsh2020application,fink2021kpi,natanian2021simplified,miao2018computational}. Since our work aims to generate comprehensive layouts that include various types of elements, we primarily compare our method with approaches capable of generating layouts containing at least both roads and buildings, including~\cite{jyothi2019layoutvae,wen2022layoutransformer,xu2021blockplanner,lin2023infinicity,xie2023citydreamer}. 

In ~\cref{table:comparison_functionality}, we provide a comparative analysis of our layout generation methods alongside recent approaches. The analysis reveals that previous methods (LayoutVAE~\cite{jyothi2019layoutvae}, LayoutTransformer~\cite{wen2022layoutransformer}, GlobalMapper~\cite{he2023globalmapper}, BlockPlanner~\cite{xu2021blockplanner}) are often constrained by the necessity for complex priors, limited by the variety of object types they can handle, restricted by the shapes of buildings they accommodate, do not scale effectively, or offer minimal user control. These limitations constrain their suitability for large-scale urban scenes featuring multiple objects with complex shapes and orientations, thus reducing their applicability to city layout generation. In the subsequent sections, we delve deeper into the analysis of Infinicity~\cite{lin2023infinicity} and CityDreamer~\cite{xie2023citydreamer}, which have similar functions as our framework and are better equipped for comprehensive city layout generation.

\subsection{Quantitave and Qualitative Analysis} 
\paragraph{Evaluation Metrics}
To quantitatively evaluate the performance of all models, we adopt two evaluation metrics: Kernel Inception Distance (KID)\cite{binkowski2018demystifying} and Fréchet Inception Distance (FID)\cite{heusel2017gans}. These two metrics are among the most commonly used standards for assessing the quality of generated outputs in generative models.

\paragraph{Block Generation Models}
We first evaluate the performance of our block generation models trained at three image scales: BG$_{512}$, BG$_{256}$, BG$_{128}$ and compared them with the state-of-the-art method InfiniCity~\cite{lin2023infinicity} and CityDreamer~\cite{xie2023citydreamer} both qualitatively and quantitatively. Since InfiniCity is not open-sourced, we compare our layout generation method with its layout generation method InfinityGAN~\cite{lin2021infinitygan}. To ensure fair comparisons, we trained InfinityGAN on our dataset in RGB space, using the same settings as outlined in the InfiniCity paper. InfinityGAN was trained for 800,000 iterations, and we selected the model with the best FID score, which occurred at epoch 231,000. In the first two rows of~\cref{fig:quality}, we present both the raw sampled outputs and the results after mapping the color of each pixel to its nearest neighbor in the predefined palette, as described in the InfiniCity paper. From the results, it is evident that InfinityGAN struggles to generate structured object layouts, such as standalone houses, and the overall quality of the outputs is suboptimal. Furthermore, it is difficult to extract meaningful semantic information from the generated results. It is also notable that the color conversion for InfinityGAN samples create undesirable artifacts near boundaries of objects (highlighted with red boxes), such as building pixels in the middle of traffic roads. In the third row we show samples from CityDreamer. From the sampled results, we observe that CityDreamer generates irregularly shaped buildings and roads, with objects lacking clear separation, particularly in the case of buildings, which often merge into large connected regions. This makes it difficult to segment individual instances. Additionally, there are numerous instances where vegetation or water is directly connected to buildings, which is uncommon in the real world; In the last three rows, we present the sampling results of CityGen's block generation model at different scales. From these results, it is evident that CityGen produces high-quality, well-defined local layouts with regular shapes and clear boundaries for buildings and roads, while capturing a variety of semantic categories. The generated outputs exhibit both diversity and quality. From the quantitative comparisons, we can tell CityGen outperforms previous methods in generating diverse and high-quality city semantic layouts. In~\cref{tab:BG_fid} we perform quantitative comparisons by calculating the FIDs and KIDs for InfinityGAN and CityDreamer to compare them with our models. We sample 10,000 samples from each model and randomly sample 10,000 real images when calculating FID and KID, we draw conclusion that our block generation models trained with images at three scales substantially outperform other methods.

\begin{table}[t]
    \centering
    \caption{\textbf{Quantitative comparison on block generation(BG)}. CityGen models substantially outperforms other methods in FID and KID (both the lower the better). BG$_{n}$ indicates the model is trained using $n^2$ images.}
    \begin{tabular}{l | r r}
         \toprule
         Method & FID $(\downarrow)$ & KID $(\downarrow)$ \\
         \midrule
         InfiniCity~\cite{lin2023infinicity}    & 80.35 &   0.0875 \\
         CityDreamer~\cite{xie2023citydreamer}    & 111.44 & 0.11  \\
         \midrule
         \textbf{CityGen~(ours)} \\
         BG$_{128}$ & 20.18 & 0.0221  \\
         BG$_{256}$ & 28.92  & 0.0290  \\
         BG$_{512}$ & 61.38 & 0.0595 \\
         \bottomrule
    \end{tabular}
    \label{tab:BG_fid}
\end{table}

\definecolor{ground}{RGB}{85, 107, 47}
\definecolor{vegetation}{RGB}{0, 255, 0}
\definecolor{building}{RGB}{255, 165, 0}
\definecolor{rail}{RGB}{255, 0, 255}
\definecolor{road}{RGB}{200, 200, 200}
\definecolor{footpath}{RGB}{255, 255, 0}
\definecolor{water}{RGB}{0, 191, 255}

\begin{figure}[t]
    \centering
    \includegraphics[width=\linewidth]{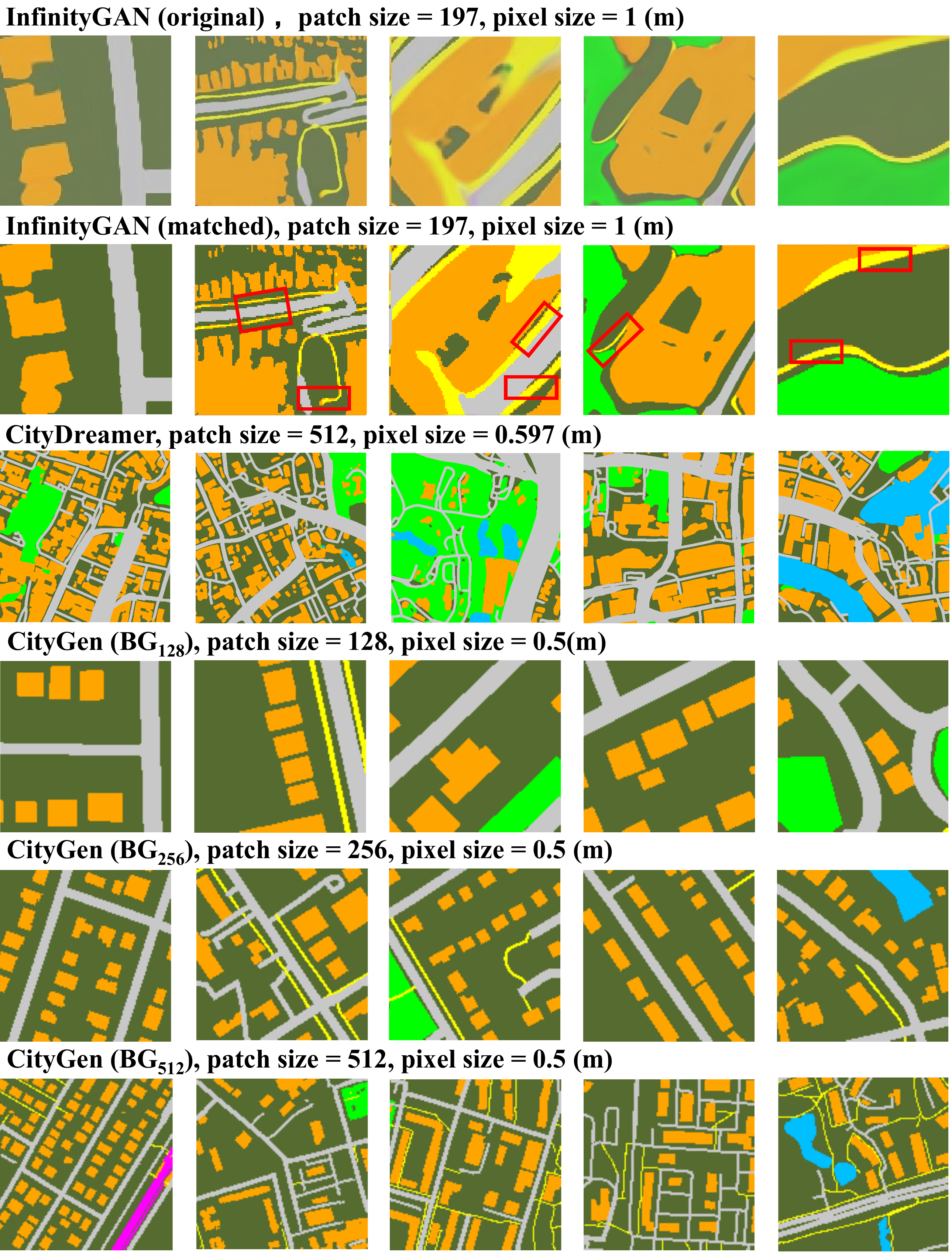}
    \parbox[c][8pt][l]{8pt}{\colorbox{ground}{}}ground\quad
    \parbox[c][8pt][l]{8pt}{\colorbox{vegetation}{}}vegetation\quad
    \parbox[c][8pt][l]{8pt}{\colorbox{building}{}}building\quad
    \parbox[c][8pt][l]{8pt}{\colorbox{rail}{}}rail\quad
    \parbox[c][8pt][l]{8pt}{\colorbox{road}{}}road\quad
    \parbox[c][8pt][l]{8pt}{\colorbox{footpath}{}}footpath\quad
    \parbox[c][8pt][l]{8pt}{\colorbox{water}{}}water\quad
    \caption{\textbf{Local Layout Samples}. Samples drawn from CityGen and Infinicity's~\cite{lin2023infinicity} block generation model InfinityGAN~\cite{lin2021infinitygan} (1). InfinityGAN's patch generation samples. (2).InfinityGAN results after nearest-neighbor color mapping. Regions in red boxes indicate artifacts caused by color mapping. (3). CityDreamer's patch generation results. (4)(5)(6).Samples from CityGen's Block Generation model trained with $128^2, 256^2, 512^2$ images, respectively. \textbf{(Note: The color palette is consistent across all figures and introduced only here.)}}
    \label{fig:quality}
\end{figure}

\begin{figure}[t]
    \centering
    \includegraphics[width=\linewidth]{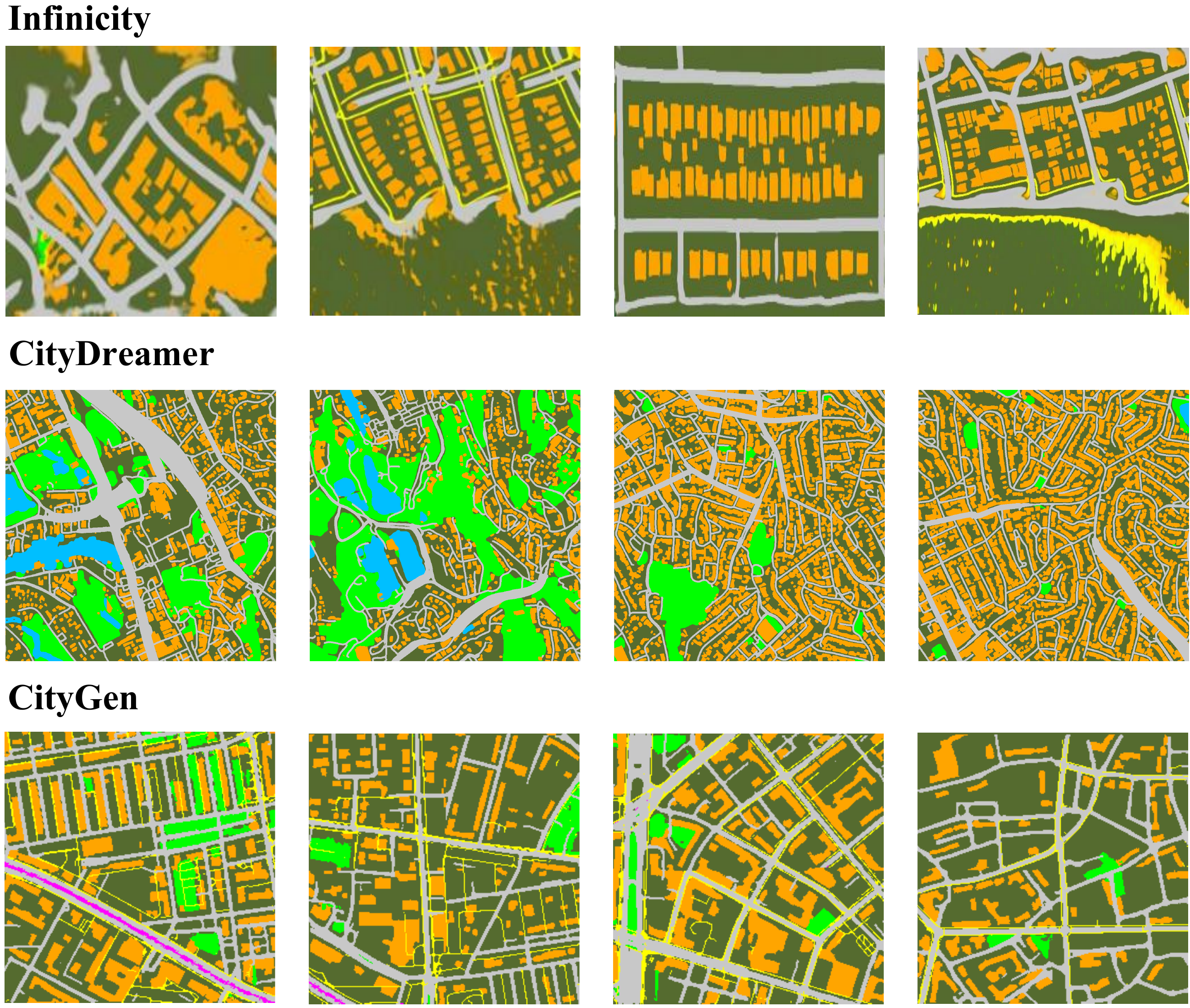}
    \caption{\textbf{Infinite Expansion Results}. Each row contains four $1024^2$ patches. (\textbf{Note: Please zoom in for details}.)}
    \label{fig:infinite_expansion}
\end{figure}

\begin{figure}[t]
    \centering
    \includegraphics[width=\linewidth]{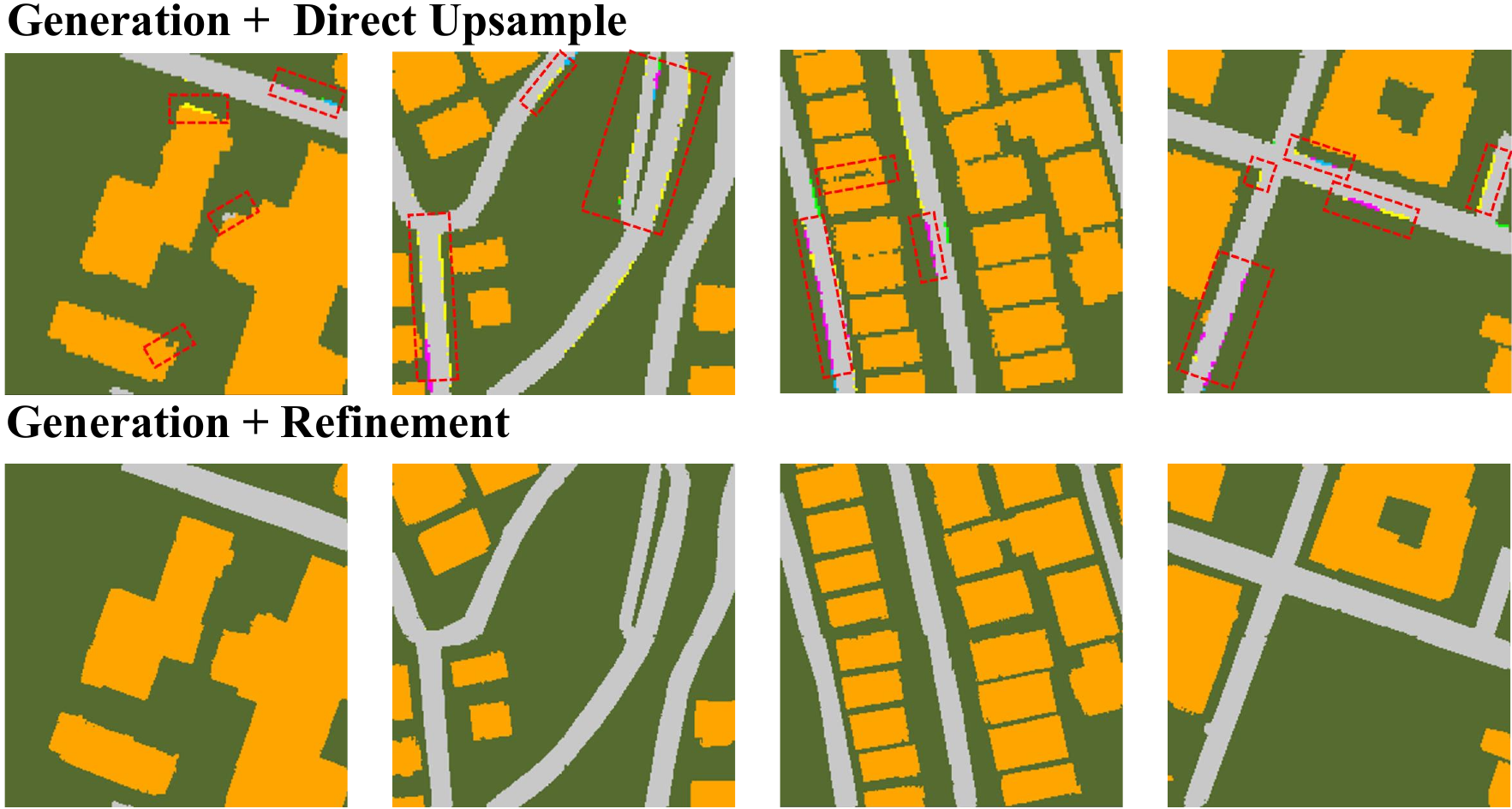}
    \caption{\textbf{Multi-scale Refinement}. (Top) Direct upsampling introduces noisy points and saw-tooth patterns. (Bottom) Multi-scale refinement removes noise and smooths object boundaries. \textbf{(Note: Please zoom in for details.)}}
    \label{fig:refine}
\end{figure}
\paragraph{Inpainting and Outpainting Models}
Our infinite expansion module and multi-scale refinement module are built on inpainting and outpainting models, so we tested both types of models. First, we quantitatively evaluated the generation capabilities of inpainting and outpainting. Since restricting the mask type and size does not fully demonstrate the model’s performance under arbitrary masks or inputs, we used random input images and random masks during the testing phase to ensure fairness. For each model, we selected 10,000 images from the test set and masked the each input image with a random ratio $\sigma \in (0.1, 0.9)$. These images were then used to compute FID and KID with other testing images, and the results are shown in~\cref{tab:In Out fid}. The results indicate that both inpainting and outpainting models effectively complete the user-specified regions, outperforming block generation models of the same scale. This is because, during the testing phase, part of the image is provided as a prior input to the inpainting or outpainting models, guiding them in reconstructing the full image more accurately.

\begin{figure}[t]
    \centering
    \includegraphics[width=\linewidth]{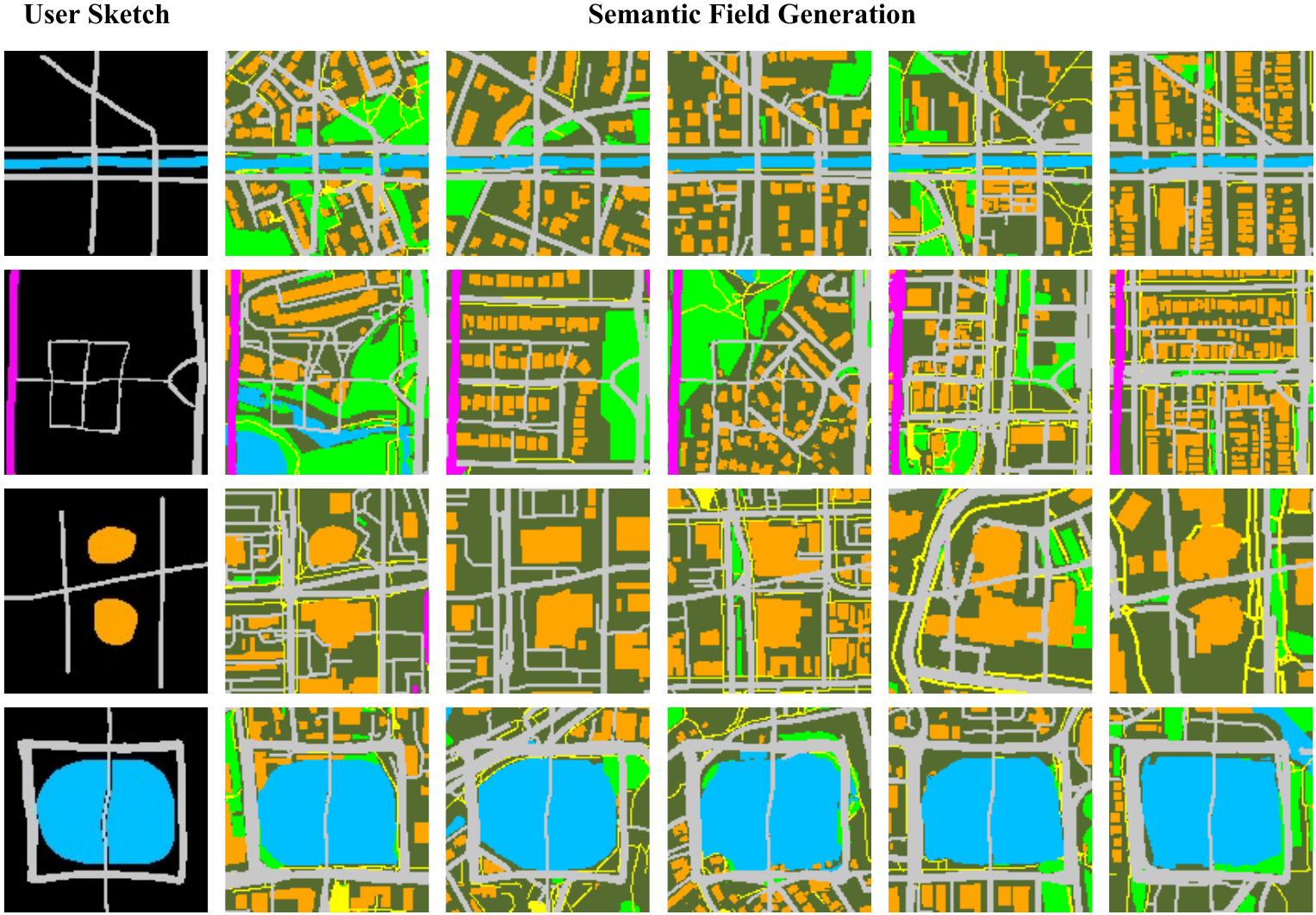}
    \caption{\textbf{User Control Results} Using user-provided sketches, we generate diverse semantic fields that meet their design requirements. Users draw directly on an empty background, after which we detect the semantics and create inpainting masks. These initial semantic layouts and masks are then input into the inpainting models to generate complete layouts.}
    \label{fig:control}
\end{figure}

\begin{figure}[t]
    \centering
    \includegraphics[width=\linewidth]{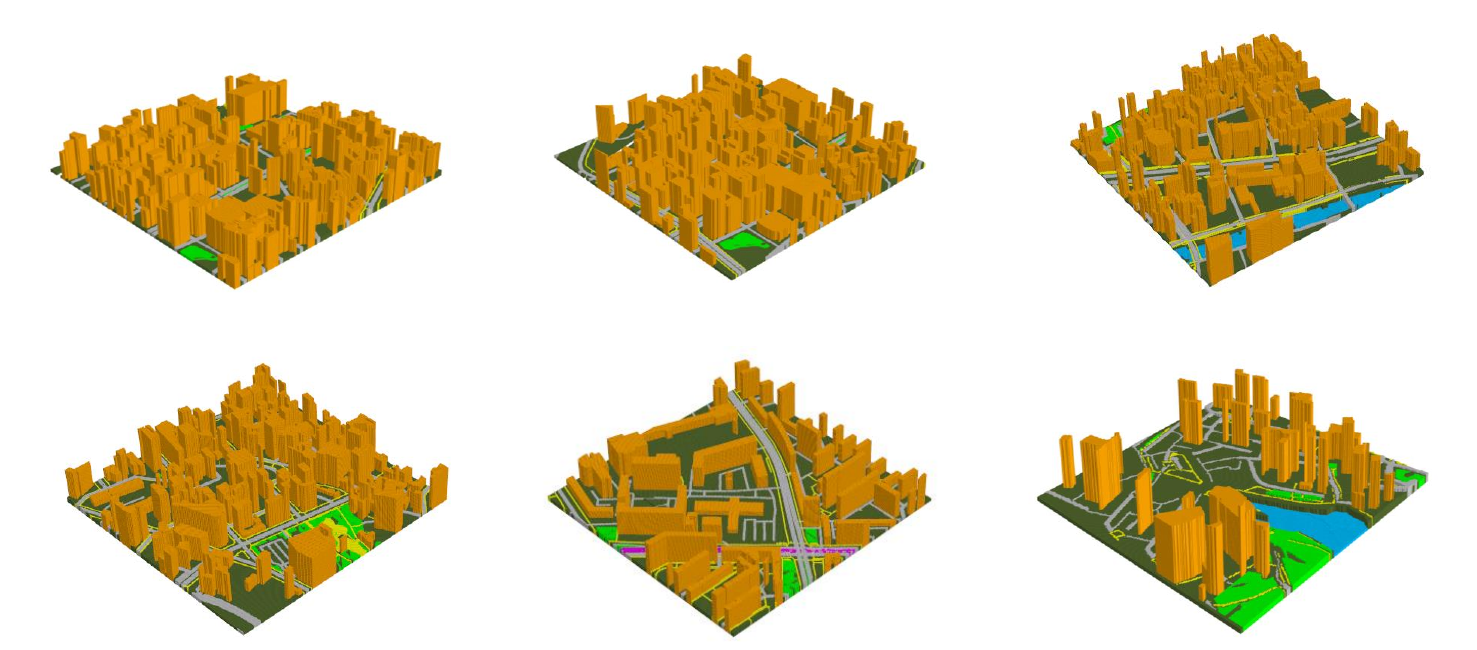}
    \caption{\textbf{Examples of Constructed 3D Layouts}. With generated semantic fields and synthesized height fields, we can construct high-quality 3D layouts. Each layout has and area of $1024 ^{2} m^{2}$}
    \label{fig:3dlayouts}
\end{figure}

To more intuitively demonstrate the effectiveness of the models, we present qualitative examples. In~\cref{fig:infinite_expansion}, we demonstrate the effectiveness of the infinite expansion module and compare the results with other methods qualitatively. The results show that Infinicity struggles to generate large, continuous areas, with many blurred shapes and uneven distribution of buildings. Additionally, the shapes of roads and buildings are inconsistent and lack coherence. In CityDreamer's results, rivers and vegetation occupy a disproportionate amount of space, and most buildings are clustered together, making it difficult to distinguish individual instances. In contrast, CityGen generates diverse, structurally rich, and consistent city layouts where instances can be clearly differentiated; In~\cref{fig:refine} we compare the results of direct upsampling versus multi-scale refinement applied to the generated images.The downsampling process during training and the upsampling process during generation inevitably introduce noisy points and saw-tooth patterns along object boundaries. As seen in the results, multi-scale refinement effectively addresses these issues, upsampling a low-resolution image into one with sharp, smooth edges and no noisy points. In~\cref{fig:control} we present examples of user-customization generation. In each row, the first column shows the user’s input sketch. During the testing phase, we allowed users to draw their desired requirements, which were then converted into an initial semantic layout, and the inpainting mask was computed. Both parts were subsequently fed into the inpainting model to generate the final completed semantic layout. The results demonstrate that CityGen is capable of generating high-quality, diverse layouts that meet user requirements; In~\cref{fig:3dlayouts} we showcase the constructed 3D layouts. For each layout, we first generate a $1024^2$ semantic field, then synthesize the corresponding height field. Finally, we combine both to construct the 3D semantic layout.

\section{Ablation Study}
In this section, we highlight the effectiveness of training block generation models in one-hot space. Using the same set of training images and techniques, we trained three block generation models in both RGB space and one-hot space for comparison. During sampling in RGB space, we mapped the RGB values of the generated samples back to their nearest neighbors in a predefined palette. We then compared the FIDs and KIDs between the RGB-trained models and those trained in one-hot space, as shown in Tab.~\ref{tab:rgb_fid}. The results demonstrate that training in one-hot space significantly improves model performance.

\begin{table}
\centering
\caption{
    \textbf{FID and KID for inpainting and outpainting models.} During training, all images are down-sampled to $128 \times 128$.
}
\begin{tabular}{l | r r} 
    \toprule
    Variant & FID $(\downarrow)$ & KID $(\downarrow)$\\
    \midrule
    Out$_{512}$& 15.38 & 0.0120   \\
    In$_{256}$ & 22.25 & 0.0171  \\
    In$_{128}$ & 27.33 & 0.0215  \\
    \bottomrule
\end{tabular}
\label{tab:In Out fid}
\end{table}

\begin{table}[t]
\centering
\caption{
    FIDs~/~KIDs($\downarrow$) for block generation models trained at all scales, comparing our one-hot space models (BG) with RGB space models (BG$_{\text{RGB}}$). The results show that training in one-hot space significantly improves performance.
}
\resizebox{1\linewidth}{!}{
\begin{tabular}{l| c c c}
\toprule
\diagbox{Model}{Scale} & 128 &  256 & 512\\
\midrule
BG & \textbf{20.18}~/~\textbf{0.0221}   & \textbf{28.92}~/~\textbf{0.0272} & \textbf{61.38}~/~\textbf{0.0595} \\
BG$_{\text{RGB}}$ & 56.15~/~0.0561 & 68.32~/~0.0656  & 85.13~/~0.0893 \\
\bottomrule
\end{tabular}
}
\label{tab:rgb_fid}
\end{table}

\section{Limitations and Future Works}
While our framework generates infinite and diverse 3D city layouts aligned with user specifications, there is room for improvement. Currently, the finest resolution achievable is 0.5 meters, and we aim to enhance this for more detailed layouts at smaller scales. Future work will focus on expanding user control by incorporating additional guidance, such as RGB images and text prompts, for more intuitive and precise generation. We also plan to include dynamic elements like moving vehicles and pedestrians, further improving CityGen's practicality and stability for a wider range of downstream tasks.

\section{Conclusion}
In this paper, we present CityGen, a novel end-to-end framework for generating infinite and controllable city layouts guided by user input. Our experiments show that CityGen outperforms existing methods in both the quality and diversity of generated layouts, while offering superior user control. This approach marks a significant advancement in urban design and modeling, offering a powerful tool for supporting downstream tasks and applications that require realistic and varied city layouts.

\clearpage

{
    \small
    \bibliographystyle{ieeenat_fullname}
    \bibliography{main}
}


\end{document}


\maketitlesupplementary
\section{Introduction}

In this supplementary material, we more visual examples to complement our paper. The supplementary material is structured as follows:

\begin{itemize}
    \item More block generation results in~\cref{sec:A}
    \item More user control results in~\cref{sec:B}
    \item Semantic and height fields pairs in~\cref{sec:C}
    \item More large-scale generation results in~\cref{sec:D}
\end{itemize}

\section{Block Generation}\label{sec:A}
\begin{figure}[h]
    \centering
    \includegraphics[width=\linewidth]{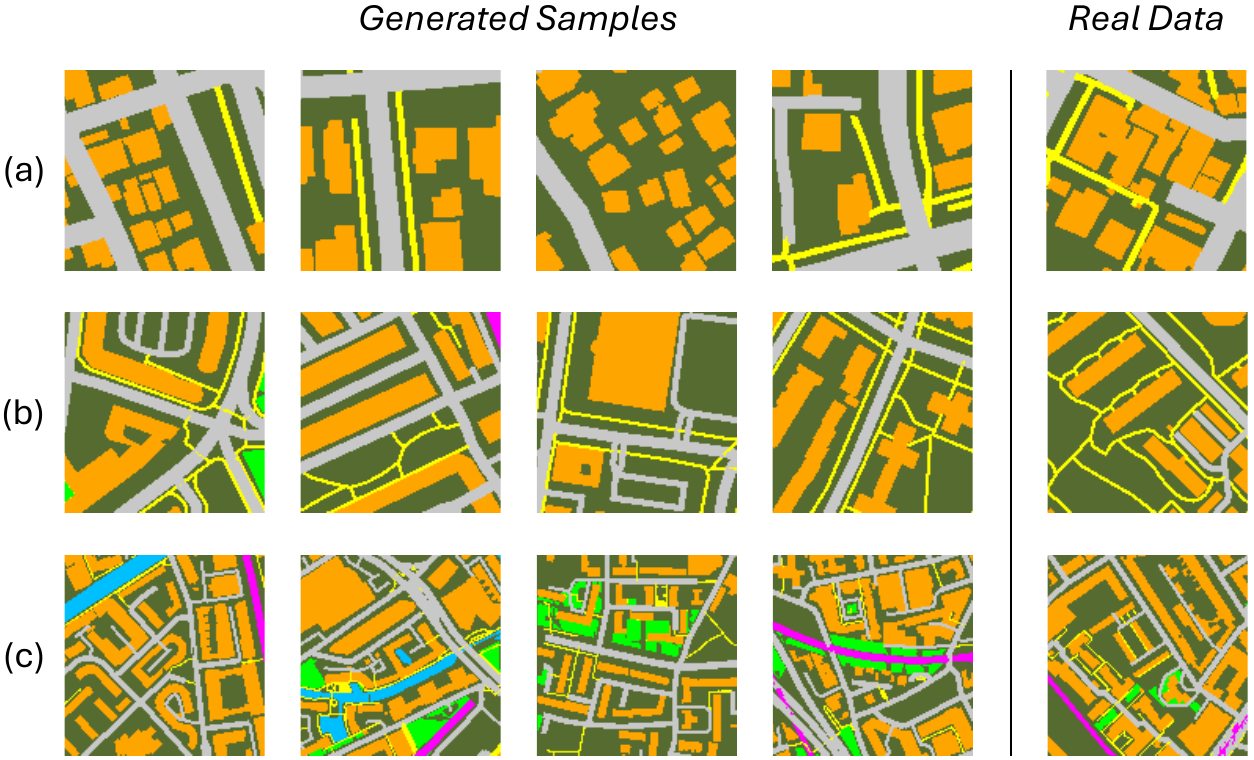}
    \caption{\textbf{Sampled Results from CityGen's Block Generation Models}. CityGen's block generation models can generate diverse and realistic 2D layouts at all scales. (a) $128^2$ (b) $256^2$ (c) $512^2$. On last column of each row we show a ground truth data in comparison to the generated results.(Class palette: \colorbox[RGB]{85, 107,47}{Terrain}, \colorbox[RGB]{0, 255,0}{Vegetation}, \colorbox[RGB]{255, 165,0}{Building}, \colorbox[RGB]{200, 200,200}{Traffic Roads}, \colorbox[RGB]{255,0, 255}{Rail}, \colorbox[RGB]{255, 255,0}{Footpath}, \colorbox[RGB]{0, 191,255}{Water}.)}
    \label{fig:Block Gen Supp}
\end{figure}

We first present more samples from CityGen's block generation models at three scales in~\cref{fig:Block Gen Supp}.

\section{User Control}\label{sec:B}
\begin{figure}[h]
    \centering
    \includegraphics[width=\linewidth]{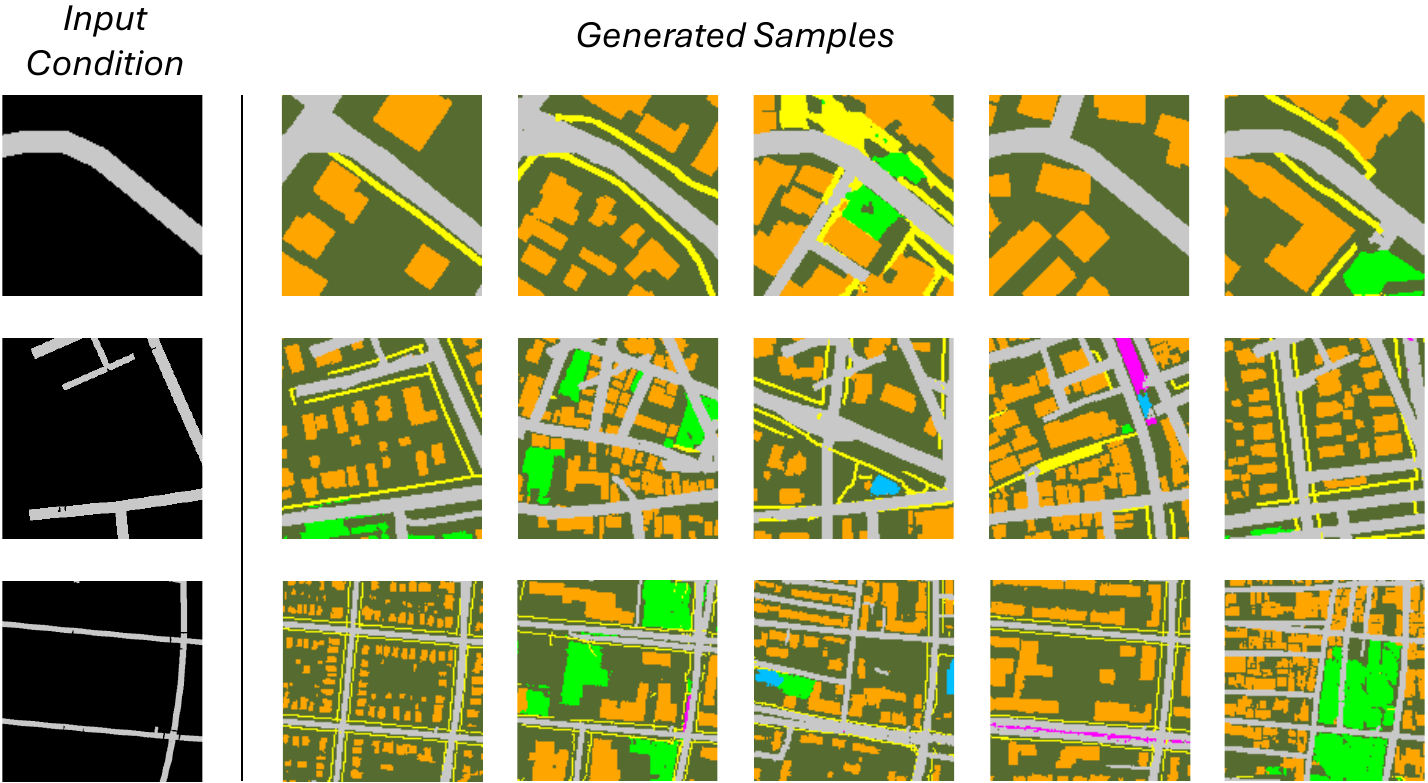}
    \caption{\textbf{Controlled Generated Samples}. Controlled Generated Samples. CityGen’s inpainting and outpainting models produce diverse, realistic results based on user input sketches at scales of 128, 256, and 512 pixels (shown in the first column).}
    \label{fig:User Control Supp}
\end{figure}

Next, we present more user control results in~\cref{fig:User Control Supp}. We input sketches of road segments as control signals.

\section{Height Field Generation}\label{sec:C}
\begin{figure}[h]
    \centering
    \includegraphics[width=\linewidth]{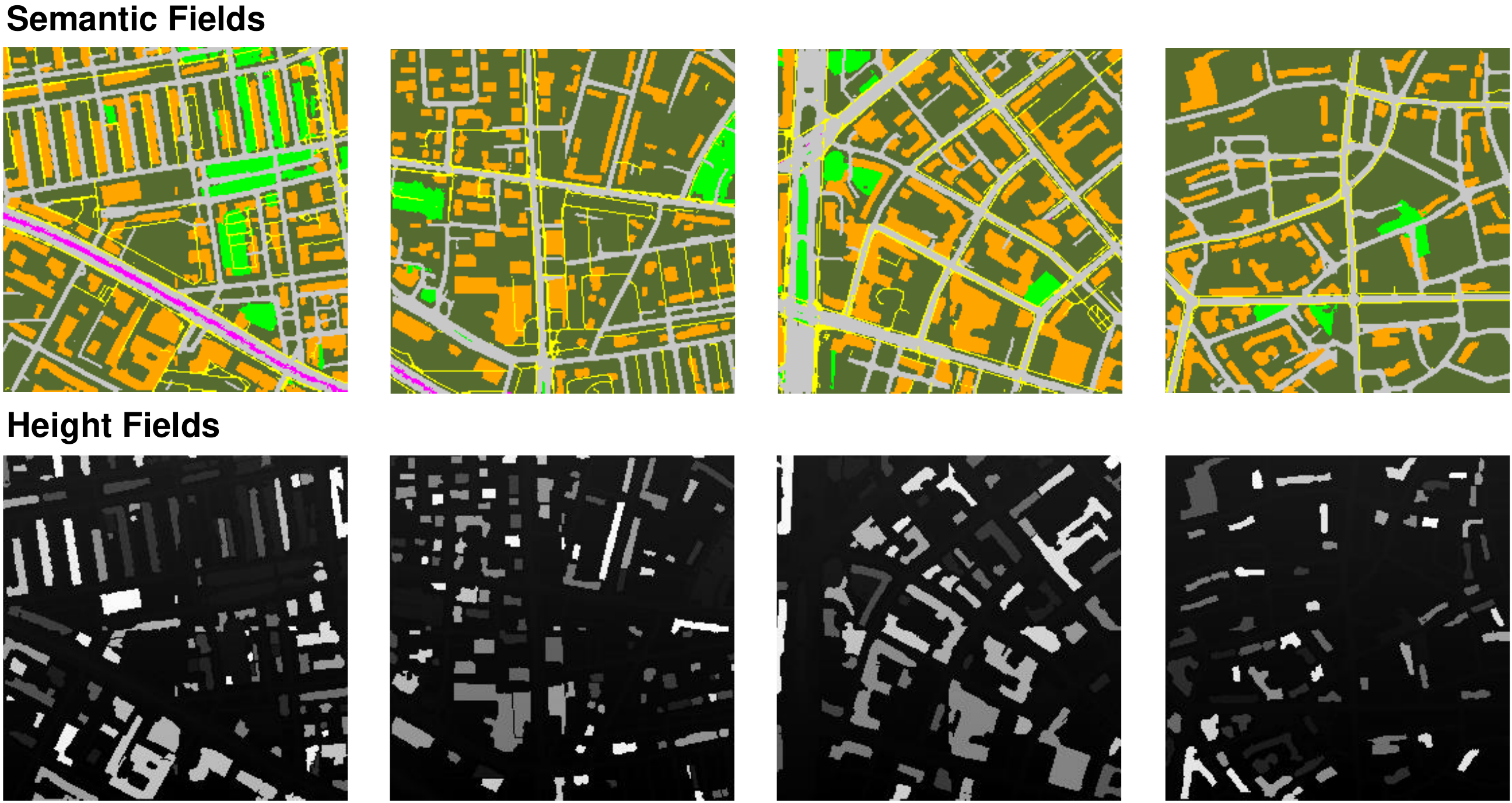}
    \caption{\textbf{Semantic Layout and Height Field Pairs}.Top: Semantic layouts of size $1024^2$; Bottom: Synthesized height fields normalized to the range of $(0, 1)$, brighter color indicate higher value.}
    \label{fig:SH supp}
\end{figure}

In this section, we show pairs of semantic layouts and height fields in~\cref{fig:SH supp}. 

\section{Large Scale Generation}\label{sec:D}
\begin{figure}[h]
    \centering
    \includegraphics[width=\linewidth]{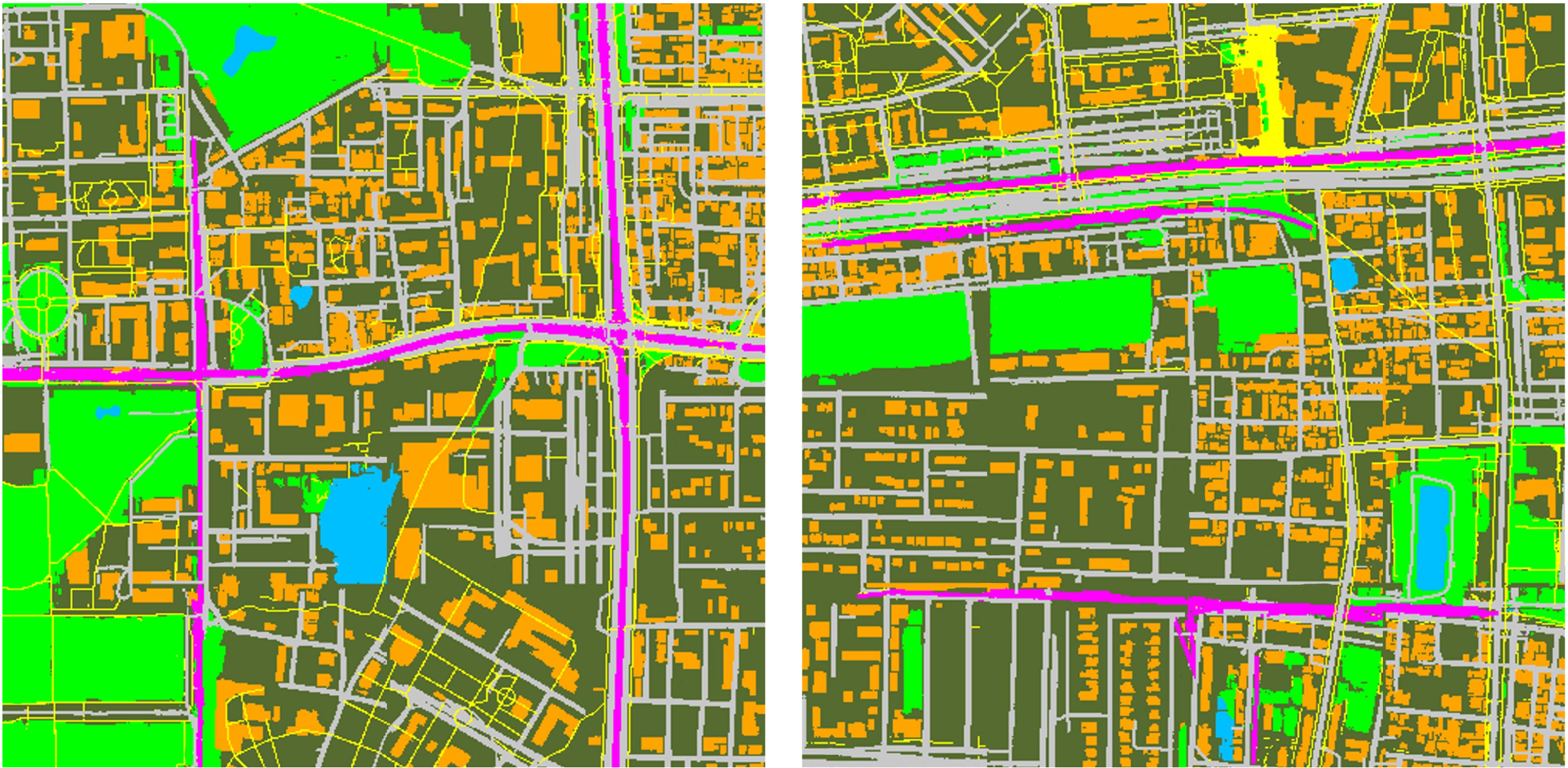}
    \caption{\textbf{Infinite expansion results}.$2048^2$ samples from CityGen's infinite expansion module.}
    \label{fig:infite WACV}
\end{figure}

In this section we present two $2048^2$ results generated using CityGen's Infinite Expansion Module in ~\cref{fig:infite WACV}. We first generate $512^2$ patches, then slide the outpainting window in various directions to generate the $2048^2$ layouts.
